\pdfoutput=1

\documentclass[11pt]{article}

\usepackage{ACL2023}

\usepackage{times}
\usepackage{latexsym}

\usepackage[T1]{fontenc}

\usepackage[utf8]{inputenc}
\usepackage{color, colortbl}
\usepackage{microtype}

\usepackage{inconsolata}
\usepackage{xcolor}         
\usepackage{xspace}
\usepackage{comment}
\usepackage{subcaption}
\usepackage{tabularx}
\usepackage{graphicx}
\usepackage{booktabs}       
\usepackage{amsfonts}   
\usepackage{textgreek}
\usepackage{amssymb}
\usepackage{amsmath}
\usepackage{newunicodechar}
\usepackage{tipa}
\usepackage[T4,T1]{fontenc}
\usepackage{siunitx}
\usepackage{pifont}

\newcommand*{\yoruba}{Yor\`ub\'a\xspace}
\newcommand*{\naija}{Nigerian-Pidgin\xspace}
\definecolor{Color}{gray}{0.8}

\newenvironment{tfour}{\fontencoding{T4}\selectfont}{}

\title{NollySenti: Leveraging Transfer Learning and Machine Translation for Nigerian Movie Sentiment Classification}

\author{Iyanuoluwa Shode$^\dagger$ \ \  \ \ \ \  David Ifeoluwa Adelani$^\ddagger$  \ \ \  \ \ \ Jing Peng$^\dagger$  \ \ \ \ \ \ Anna Feldman$^\dagger$ \\
         $^\dagger$Montclair State University, USA, and  $^\ddagger$University College London, United Kingdom \\
         \texttt{\{shodei1,pengj,feldmana\}@montclair.edu},  \ \  \ \texttt{d.adelani@ucl.ac.uk} \\}

\begin{document}
\maketitle
\begin{abstract}
Africa has over 2000 indigenous languages but they are under-represented in NLP research due to lack of datasets. 
In recent years, there have been progress in developing labelled corpora for African languages. However, they are often available in a single domain and may not generalize to other domains. 
%
In this paper, we focus on the task of sentiment classification for cross-domain adaptation. We create a new dataset, NollySenti---based on the Nollywood movie reviews for five languages widely spoken in Nigeria (English, Hausa, Igbo, \naija, and \yoruba). We provide an extensive empirical evaluation using classical machine learning methods and pre-trained language models. 
%
Leveraging transfer learning, we compare the performance of cross-domain adaptation from Twitter domain, 
and cross-lingual adaptation from English language. Our evaluation shows that transfer from English in the same target domain leads to more than 5\% improvement in accuracy compared to transfer from Twitter in the same language. 
To further mitigate the domain difference, we leverage machine translation (MT) from English to other Nigerian languages, which leads to a further improvement of 7\% over cross-lingual evaluation. 
While MT to low-resource languages are often of low quality, through human evaluation, we show that most of the translated sentences preserve the sentiment of the original English reviews. 






\end{abstract}

\section{Introduction}

Nigeria is the sixth most populous country in the world\footnote{\url{https://www.census.gov/popclock/print.php?component=counter}} and the most populous in Africa with over 500 languages~\citep{ethnologue}. These languages are spoken by millions of speakers, and the four most spoken indigenous languages (Hausa, Igbo, \naija (Naija), and \yoruba) have more than 25 million speakers but they are still under-represented in NLP research~\citep{adebara-abdul-mageed-2022-towards,van-esch-etal-2022-writing}. The development of NLP for Nigerian languages and other African languages is often limited by a lack of labelled datasets
~\cite{adelani-etal-2021-masakhaner,joshi-etal-2020-state}. While there have been some progress in recent years~\cite{eiselen-2016-government,adelani-etal-2022-masakhaner,team2022NoLL,Muhammad2023AfriSentiAT,Adelani2023MasakhaNEWSNT}, 
most benchmark datasets for African languages are only available in a single domain, and may not 
transfer well to other target domains of interest~\cite{adelani-etal-2021-effect}. 

One of the most popular NLP tasks is sentiment analysis. 
In many high-resource languages like English, sentiment analysis datasets are available across several domains like social media posts/tweets~\cite{rosenthal-etal-2017-semeval}, product reviews~\cite{Zhang_yelp,He_amazonreviews} and movie reviews~\cite{pang-lee-2005-seeing,maas-etal-2011-learning}. However, for Nigerian languages, the only available dataset is NaijaSenti~\cite{muhammad-etal-2022-naijasenti} - a Twitter sentiment classification dataset for four most-spoken Nigerian languages. It is unclear how it transfers to other domains. 

In this paper, we focus on the task of sentiment classification for cross-domain adaptation. We create the first sentiment classification dataset for Nollywood movie reviews known as \textbf{NollySenti} --- a  dataset for five widely spoken Nigerian languages (English, Hausa, Igbo, \naija, and \yoruba). Nollywood is the home for Nigerian movies that depict the Nigerian people and reflect the diversities across Nigerian cultures. Our choice of this domain is because Nollywood is the second-largest movie and film industry in the world by annual output\footnote{\url{https://www.masterclass.com/articles/nollywood-new-nigerian-cinema-explained}}, and the availability of Nollywood reviews on several online websites. However, most of these online reviews are only in English. To cover more languages, we asked  professional translators to translate about 1,000-1,500 reviews from English to four Nigerian languages, similar to \citet{winata-etal-2023-nusax}. Thus, \textbf{NollySenti} is a \textbf{parallel multilingual sentiment corpus} for five Nigerian languages that can be used for both \textit{sentiment classification} and \textit{evaluation of machine translation (MT) models} in the user-generated texts domain --- which is often scarce for low-resource languages. 

Additionally, we provide several supervised and transfer learning experiments using classical machine learning methods and pre-trained language models. By leveraging transfer learning, we compare the performance of cross-domain adaptation from the Twitter domain to the Movie domain, and cross-lingual adaptation from English language. Our evaluation shows that transfer from English in the same target domain leads to more than 5\% improvement in accuracy compared to transfer from the Twitter domain in the same target language. To further mitigate the domain difference, we leverage MT from English to other Nigerian languages, which leads to a further improvement of 7\% over cross-lingual evaluation. While MT to low-resource languages are often of low quality, through human evaluation, we show that most of the translated sentences preserve the sentiment in the original English reviews.  For reproducibility, we have released our datasets and code on Github\footnote{\url{https://github.com/IyanuSh/NollySenti}}.


\section{Related Work}
\paragraph{African sentiment datasets} There are only a few sentiment classification datasets for African languages such as Amharic dataset~\citep{yimam-etal-2020-exploring}, and NaijaSenti~\citep{muhammad-etal-2022-naijasenti}--- for Hausa, Igbo, \naija, and \yoruba. Recently, \citet{Muhammad2023AfriSentiAT} expanded the sentiment classification dataset to 14 African languages. However, all these datasets belong to the social media or Twitter domain. In this work, we create a new dataset for the Movie domain based on human translation from English to Nigerian languages, similar to the NusaX parallel sentiment corpus for 10 Indonesia languages~\citep{winata-etal-2023-nusax}. 

\paragraph{MT for sentiment classification} In the absence of training data, MT models can be used to translate texts from a high-resource language like English to other languages, but they often introduce errors that may lead to poor performance~\citep{refaee-rieser-2015-benchmarking,poncelas-etal-2020-impact}. However, they do have a lot of potentials especially when translating between high-resource languages like European languages, especially when combined with English~\citep{balahur-turchi-2012-multilingual,balahur-turchi-2013-improving}. In this paper, we extend MT for sentiment classification to four low-resource Nigerian languages. This paper is an extension of the YOSM paper~\cite{shode2022yosm} -- A \yoruba movie sentiment corpus. 
\section{Languages and Data}
\label{sec:focus_languages}

\subsection{Focus Languages}
We focus on four Nigerian languages from three different language families spoken by 30M-120M. 

\paragraph{Hausa} belongs to  the Afro-Asiatic/Chadic language family with over 77 million speakers~\citep{ethnologue}. It is a native to Nigeria, Niger, Chad, Cameroon, Benin, Ghana, Togo, and  Sudan. However, the significant population for the language reside in northern Nigeria. 
Hausa is an agglutinative language in terms of morphology and tonal with two tones --- low and high. It is written with two major scripts: Ajami (an Arabic-based script) and Boko script (based on Latin script) --- the most widely used. The Boko script make use of all the Latin letters except for ``p,q,v, and x'' including the following additional letters ``\texthtb, \texthtd, \texthtk, \begin{tfour}\m{y}\end{tfour}, kw, {\texthtk}w, gw, ky, {\texthtk}y, gy, sh, and ts''. 

\paragraph{Igbo} belongs to the Volta–Niger sub-group of the Niger-Congo language family with over 31 million speakers~\cite{ethnologue}. It is native language to South-Eastern Nigeria, but also spoken in Cameroon and Equatorial Guinea in Central Africa. 
Igbo is an agglutinative language in terms of its sentence morphology and tonal with two tones - high and low.  The language utilizes 34 Latin letters excluding ``c,q and x'', however, it includes additional letters ``ch, gb, gh, gw, kp, kw, nw, ny, {\d{o}}, $\dot{o}$, {\d{u}} and sh''. 

\paragraph{Nigerian-Pidgin aka Naija} is from the English Creole Atlantic Krio language family with over 4 million native speakers and 116 million people second language speakers. It is a broken version of Nigerian English that is also a creole because it is used as a first language in certain ethnic communities~\citep{mazzoli_naija}. It serves as a common language for all as it facilitates communication between several ethnicities. Naija has 26 letters similar to English with an analytical sentence morphology.

\paragraph{\yoruba} belongs to the Volta–Niger branch of the Niger-Congo language family with over 50 million speakers ~\cite{ethnologue} thus making it the third most spoken indigenous African language. \yoruba is native to South-Western Nigeria, Benin and Togo, and widely spoken across West Africa and Southern America like Sierra Leone, Côte d'Ivoire, The Gambia, Cuba, Brazil, and some Caribbean countries. \yoruba is an isolating language in terms of its sentence morphology and tonal with three lexical tones - high, mid and low - that are usually marked by diacritics which are used on syllabic nasals and vowels. \yoruba orthography comprises 25 Latin letters which excludes ``c, q, v, x, and z'' but  includes additional letters ``gb, {\d{e}}, {\d{s}} and {\d{o}}''.

\subsection{NollySenti creation}

Unlike Hollywood movies that are heavily reviewed with hundreds of thousands of reviews all over the internet, there are fewer reviews about Nigerian movies despite their popularity. Furthermore, there is no online platform dedicated to writing or collecting movie reviews written in the four indigenous Nigerian languages. We only found reviews in English. 
Here, we describe the data source for the Nollywood reviews and how we created parallel review datasets for four Nigerian languages. 

\subsubsection{Data Source}
\autoref{tab:data_source} shows the data source for the NollySenti review dataset. We collected 1,018 positive reviews (POS) and 882 negative reviews (NEG). These reviews were accompanied with ratings and were sourced from three popular online movie review platforms 
- \textbf{IMDB}, \textbf{Rotten Tomatoes} and, \textbf{Letterboxd}. We also collected reviews and ratings from four Nigerian websites like \textbf{Cinemapointer},  \textbf{Nollyrated}.   
Our annotation focused on the classification of the reviews based on the ratings that the movie reviewer gave the movie. We used a rating scale to classify the POS or NEG reviews and defined ratings between 0-4 to be in the NEG category and 7-10 as POS.

\begin{table*}[t]
 \begin{center}
 \scalebox{0.9}{
 \footnotesize
  \begin{tabular}{l|r|r|rrrrrr}
    \toprule
	& \textbf{No. } & \textbf{Ave. Length} & \multicolumn{5}{c}{\textbf{Data source}} \\
    \textbf{Sentiment} & \textbf{Reviews} & \textbf{(No. words)} & \textbf{IMDB} & \textbf{Rotten Tomatoes} & \textbf{LetterBoxd} & \textbf{Cinemapoint} & \textbf{Nollyrated} & \textbf{Others} \\
    \midrule
    positive & 1018 & 35.0 & 493 & 107 & 81 & 154 & 181 &  2\\
    negative & 882 & 20.7 & 292 & 140 & 101 & 269 & 74 & 6 \\
    \midrule
    Total & 1900 & -- & 785 & 247 & 182 & 423 & 255 & 8  \\
    
    \bottomrule
  \end{tabular}
  }
  \vspace{-2mm}
  \caption{\textbf{Data source, number of movie reviews per source, and average length of reviews }}
  \label{tab:data_source}
  \end{center}
  \vspace{-5mm}
\end{table*}

\begin{table}[!ht]
    \centering
    \scalebox{0.90}{
    \begin{tabular}{lrrr|rr}
    \toprule
         & \multicolumn{3}{c}{\textbf{Train}} & \textbf{Dev} & \textbf{Test} \\
         \textbf{Language} & \textbf{pos} & \textbf{neg} & \textbf{all} & \textbf{all} & \textbf{all} \\
        \midrule
        English (eng) & 1018 & 882 & 1300 & 100 & 500 \\ 
        Hausa (hau) & 200 & 210 & 410 & 100 & 500 \\ 
        Igbo (ibo) & 200 & 210 & 410 & 100 & 500 \\ 
        Naija (pcm) & 200 & 210 & 410 & 100 & 500 \\ 
        \yoruba (yor) & 450 & 450 & 900 & 100 & 500 \\ 
        \bottomrule
    \end{tabular}
    }
    \vspace{-2mm}
     \caption{\textbf{Dataset split.} The DEV and TEST split have equal number samples in positive and negative classes}
     \label{tab:data_split}
     \vspace{-5mm}
\end{table}
\subsubsection{Human Translation}
We hire professional translators in Nigeria and ask them to translate 1,010 reviews randomly chosen from the 1,900 English reviews. Thus, we have a parallel review dataset in English and other Nigerian languages and their corresponding ratings. For quality control, we ask a native speaker per language to manually verify the quality of over 100 randomly selected translated sentences, and we confirm that they are good translations, and they are not output of Google Translate (GT).\footnote{Easy to verify for languages with diacritics like \yoruba since GT ignores diacritics. GT does not support Naija }
All translators were properly remunerated according to the country rate\footnote{\$450 per language except for \texttt{yor} with more reviews }. In total, we translated 500 POS reviews and 510 NEG reviews. We decided to add 10 more NEG reviews since they are often shorter -- like one word e.g. ("disappointing").  


\section{Experimental Setup}
\paragraph{Data Split} \autoref{tab:data_split} shows the data split into \textbf{Train}, \textbf{Dev} and \textbf{Test} splits. They are 410/100/500 for \texttt{hau}, \texttt{ibo} and \texttt{pcm}. To further experiment with the benefit of adding more reviews, we 
translate 490 more reviews for \texttt{yor}. The ratio split for \texttt{yor} is 900/100/500, while for \texttt{eng} is 1,300/100/500. We make use of the same reviews for \textbf{Dev} and \textbf{Test} for all languages. For our experiments of transfer learning and machine translation, we make use of all the training reviews for English (i.e 1,300). We make use of a larger test set (i.e. 500 reviews) for \texttt{hau}, \texttt{ibo} and \texttt{pcm} because the focus of our analysis is on zero-shot transfer, we followed similar data split as XCOPA~\cite{ponti-etal-2020-xcopa}, COPA-HR~\cite{Ljubei2021BERTiT} and NusaX datasets.  The small training examples used in NollySenti provides an opportunity for researchers to develop more data efficient cross-lingual methods for under-resourced languages since this is a more realistic scenario.



\subsection{Baseline Models}
Here, we train sentiment models using classical machine learning models like Logistic regression and Support Vector Machine (SVM) and \textit{fine-tune} several pre-trained language models (PLMs). Unlike classical ML methods, PLMs can be used for cross-lingual transfer and often achieve better results~\citep{devlin-etal-2019-bert, winata-etal-2023-nusax}. We fine-tune the following PLMs: mBERT~\citep{devlin-etal-2019-bert}, XLM-R~\citep{conneau-etal-2020-unsupervised}, mDeBERTaV3~\citep{He2021DeBERTaV3ID}, AfriBERTa~\citep{ogueji-etal-2021-small}, and AfroXLMR~\cite{alabi-etal-2022-adapting}. The last two PLMs have been pre-trained or adapted to all the focus languages. For XLM-R and AfroXLMR, we make use of the base versions. The classical ML methods were implemented using Scikit-Learn~\cite{scikit-learn}. \autoref{sec:parameters} provides more details. 

\subsection{Zero-shot Adaptation}
\subsubsection{Transfer Learning}
\paragraph{Cross-domain adaptation} We train on the Twitter domain and perform cross-domain adaptation to the Nollywood movie domain. We make use of the NaijaSenti dataset for training. The datasets consist of between 12k-19k tweets for each of the Nigerian languages, 30 folds larger than our dataset. 

\paragraph{Cross-lingual adaptation} We train on two English datasets: (1) IMDB~\cite{maas-etal-2011-learning} -- with 25,000 reviews and (2) NollySenti English with 1,300 reviews. The resulting models are evaluated on the test set of the remaining Nigerian languages. 

\subsubsection{Machine Translation} 
Lastly, we make use of MT to mitigate the domain difference. We make use of  NLLB~\citep{team2022NoLL}\footnote{\url{https://huggingface.co/facebook/nllb-200-distilled-600M}} for \texttt{hau}, \texttt{ibo}, and \texttt{yor} languages. NLLB is a multilingual MT trained on 200 languages and dialects. It includes the three Nigerian languages except for \naija. For \naija, we make use of a pre-trained \texttt{eng$\rightarrow$pcm} MT model by \citet{adelani-etal-2022-thousand} -- trained on both religious and news domain.

\begin{table*}[!ht]
    \centering
    \resizebox{\textwidth}{!}{
    \begin{tabular}{llrr|rrr|rr|r|r}
    \toprule
         & \textbf{Parameter}& \multicolumn{2}{c}{\textbf{eng}} & \textbf{hau} & \textbf{ibo} & \textbf{pcm} & \multicolumn{2}{c}{\textbf{yor}} & &  \\
        Model & \textbf{size}& \texttt{N=410} & \texttt{N=1300} & \texttt{N=410} & \texttt{N=410} & \texttt{N=410} & \texttt{N=410} & \texttt{N=900} & \textbf{avg} & \textbf{avg (excl. eng)} \\
        \midrule
        LogisticReg & $<$20K & 79.2 & 84.2 & 78.8 & 81.8 & 83.4 & 78.8 & 80.1 & $81.0_{\pm 0.2}$ & $80.8_{\pm 0.2}$\\
        SVM & $<$20K & 79.0 & 85.2 & 79.0 & 80.6 & 83.6 & 79.7 & 81.9 & $81.3_{\pm 0.6}$ & $81.0_{\pm 0.6}$\\
        \midrule
        mBERT & 172M & 90.3 & 92.6 & 80.0 & 82.4 & 89.1 & 84.8 & 87.8 & $87.0_{\pm 0.5}$ & $85.2_{\pm 0.5}$ \\ 
        XLM-R-base & 270M & 93.2 & \textbf{94.1} & 76.8 & 83.6 & 90.8 & 83.9 & 86.0 & $86.9_{\pm 0.5}$ &  $84.2_{\pm 0.5}$ \\ 
        mDeBERTaV3 & 276M & \textbf{94.2} & \textbf{95.1} & 83.7 & 87.1 & \textbf{91.8} & 82.2 & 87.4 & $\mathbf{88.8_{\pm 0.5}}$ & $86.4_{\pm 0.5}$ \\ 
        AfriBERTa-large & 126M & 86.2 & 89.5 & \textbf{87.2} & \textbf{88.4} & 88.3 & \textbf{85.9} & \textbf{90.9} & $88.1_{\pm0.3}$ & $\mathbf{88.1_{\pm 0.3}}$ \\ 
        AfroXLMR-base & 270M & 92.3 & 94.1 & 84.2 & 85.6 & 91.0 & 83.8 & 88.4 & $88.5_{\pm0.8}$ & $86.6_{\pm 0.8}$\\
        \bottomrule
    \end{tabular}
    }
    \vspace{-2mm}
     \caption{\textbf{Baseline result using classical machine learning and pre-trained language models.} We make use of the number of training examples, $N=410, 900, $ and $1300$.  We report accuracy. Average performed over 5 runs.}
     \label{tab:baseline_result}
     \vspace{-2mm}
\end{table*}

\begin{table}[!ht]
    \centering
    \resizebox{\columnwidth}{!}{%
    \begin{tabular}{lrrrr|r}
    \toprule
         & \textbf{hau} & \textbf{ibo} & \textbf{pcm} & \textbf{yor} & 
         \textbf{ave} \\
        \midrule
        Twitter (lang) & 76.7 & 78.4 & 74.1 & 66.0 & $73.8_{\pm 0.6}$ \\ 
        IMDB (eng) & 71.3 & 71.2 & 84.0 & 66.4 & $73.2_{\pm 2.2}$ \\ 
        NollySenti (eng) & \underline{80.2}  & \underline{78.9} & \underline{86.2} & \underline{72.8} & $\underline{79.5_{\pm 2.9}}$ \\ 

        \midrule
        \multicolumn{2}{l}{\textbf{machine translation (en $\rightarrow$ lang})} \\
        IMDB (lang, \texttt{N=25k}) & 86.8 & 83.8 & 86.8 & 82.0 & $83.0_{\pm 1.0}$ \\ 
        NollySenti (lang, \texttt{N=410}) & 84.0 & 86.3 & 81.2 & 83.0 & $83.6_{\pm 0.6}$ \\ 
        NollySenti (lang) & 88.3 & 86.5 & 87.0 & \textbf{84.0} & $86.4_{\pm 0.2}$ \\ 
        NollySenti (eng+lang) & \textbf{89.5} & \textbf{86.8} & \textbf{87.2} & 83.8 & $\mathbf{86.8_{\pm 0.3}}$ \\
        \midrule
        \rowcolor{Color}
       Supervised & 87.2 & 88.4 & 88.3 & 90.9 &  $88.7_{\pm 0.3}$ \\
        \bottomrule
    \end{tabular}
    }
    \vspace{-2mm}
     \caption{\textbf{Zero-shot scenario using AfriBERTa-large:} cross-domain (Twitter -> Movie),  cross-lingual experiments (eng -> lang) and review generation using machine translation (Meta's NLLB and MAFAND~\cite{adelani-etal-2022-thousand} \texttt{eng$\rightarrow$pcm} model)}
     \label{tab:zero_shot_eval}
\end{table}

\begin{table}[!ht]
    \centering
    \resizebox{\columnwidth}{!}{%
    \begin{tabular}{lrr|rr}
    \toprule
         \textbf{Lang.} & \textbf{BLEU} & \textbf{CHRF} & \textbf{Adequacy} & \textbf{sentiment preservation} \\
        \midrule
        hau & 13.6 & 40.8 & 4.4 & 92.0\%\\ 
        ibo & 9.8 & 33.4 & 3.8 &  92.0\%\\ 
        pcm & 26.4 & 53.0 & 4.6 & 96.0\% \\ 
        yor & 3.53 & 16.9 & 4.0 &  89.5\% \\ 
        \bottomrule
    \end{tabular}
    }
    \vspace{-2mm}
     \caption{\textbf{Automatic} (\texttt{N=410}) and \textbf{Human evaluation} (\texttt{N=100}) of the MT generated reviews from TRAIN split.}

     \vspace{-2mm}
     \label{tab:automatic_metrics}
\end{table}

\section{Results}
\subsection{Baseline Results}
\autoref{tab:baseline_result} provides the baseline results using both logistic regression, SVM, and several PLMs. All baselines on average have over 80\% accuracy. However, in all settings (i.e. all languages and number of training samples, \texttt{N=400,900}, and \texttt{1300}), PLMs exceed the performance of classical machine learning methods by over $5-7\%$. In general, we find Africa-centric PLMs (AfriBERTa-large and AfroXLMR-base) have better accuracy than massively multilingual PLMs pre-trained on around 100 languages. Overall, AfriBERTa achieves the best result on average, but slightly worse for English and Nigerian-Pidgin (an English-based creole language) since it has not been pre-trained on the English language. 

\subsection{Zero-shot Evaluation Results}
We make use of AfriBERTa for the zero-shot evaluation since it gave the best result in \autoref{tab:baseline_result} (see avg. excl. \texttt{eng}). \autoref{tab:zero_shot_eval} shows the zero-shot evaluation. 
\paragraph{Performance of Cross-domain adaptation} We obtained an impressive zero-shot result by evaluating a Twitter sentiment model (i.e. \texttt{Twitter (lang)}) on movie review ($73.8$ on average). All have over $70$ except for \texttt{yor}. 

\paragraph{Performance Cross-lingual adaptation} We evaluated two sentiment models, trained on either \texttt{imdb} or \texttt{NollySenti (eng)} English reviews. Our result shows that the adaptation of \texttt{imdb} has similar performance as the cross-domain adaptation, while the \texttt{NollySenti (eng)} exceeded the performance by over $+6\%$. The \texttt{imdb} model (i.e \texttt{imdb (eng)}) was probably worse despite the large training size due to a slight domain difference between Hollywood reviews and Nollywood reviews --- may be due to writing style and slight vocabulary difference among English dialects~\cite{blodgett-etal-2016-demographic}. An example of a review with multiple indigenous named entities including a \texttt{NEG} sentiment is \textit{``\textbf{'Gbarada'} is a typical \textbf{\underline{Idumota}} 'Yoruba film' with all the craziness that come with that sub-section of Nollywood. 
''} that may not frequently occur in Hollywood reviews. Another  observation is that the performance of \texttt{pcm} was unsurprisingly good for both setups ($84.0$ to $86.2$) because it is an English-based creole.


\paragraph{Machine Translation improves adaptation} To mitigate the domain difference, we found that by automatically translating \texttt{N=410} reviews using a pre-trained MT model improved the average zero-shot performance by over $+4\%$. With additional machine translated reviews (\texttt{N=1300}), the average performance improved further by $+3\%$. Combining all translated sentences with English reviews does not seem to help. Our result is quite competitive to the supervised baseline ($-1.9\%)$. As an additional experiment, we make use of MT to translate 25k IMDB reviews, the result was slightly worse than NollySenti (lang). This further confirms the slight domain difference in the two datasets.  


\paragraph{Sentiment is often preserved in MT translated reviews} \autoref{tab:automatic_metrics} shows that despite the low BLEU score $(<15)$ for \texttt{hau}, \texttt{ibo} and \texttt{yor}, native speakers (two per language) of these languages rated the machine translated reviews in terms of content preservation or adequacy to be much better than average ($3.8$ to $4.6$) for all languages on a Likert scale of 1-5. Not only does the MT models preserve content, native speakers also rated their output to preserve more sentiment (i.e. achieving at least of 90\%) even for some translated texts with low adequacy ratings. \autoref{sec:human_eval} provides more details on the human evaluation and  examples. 

\section{Conclusion}
In this paper, we focus on the task of sentiment classification for cross-domain adaptation. We developed a new dataset, \textbf{NollySenti} for five Nigerian languages. Our results show the potential of both transfer learning and MT for developing sentiment classification models for low-resource languages. As a future work, we would like to extend the creation of movie sentiment corpus to more African languages.

\section*{Limitations}
One of the limitations of our work is that we require some form of good performance of machine translation models to generate synthetic reviews for sentiment classification. While our approach seems to work well for some low-resource languages like \texttt{yor} with BLEU score of $3.53$, it may not generalize to other sequence classification tasks like question answering where translation errors may be more critical. 

\section*{Ethics Statement}
We believe our work will benefit the speakers of the languages under study and the Nollywood industry. We look forward to how this dataset can be used to improve the processes of the Nollywood industry and provide data analytics on movies. 

We acknowledge that there maybe some bias introduced due to manually translating the dataset from English, but we do not see any potential harm in releasing this dataset. While the texts were crawled online, they do not contain personal identifying information.

\section*{Acknowledgements}
 This material is partly based upon work supported by the National Science Foundation under Grant Numbers: 2226006, 1828199, and 1704113. We appreciate Aremu Anuoluwapo for coordinating and verifying the translation of the reviews to the Nigerian languages. We appreciate the collective efforts of the following people: Bolutife Kusimo, Oluwasijibomi Owoka, Oluchukwu Igbokwe, Boluwatife Omoshalewa Adelua, Chidinma Adimekwe, Edward Agbakoba, Ifeoluwa Shode, Mola Oyindamola, Godwin-Enwere Jefus, Emmanuel Adeyemi, Adeyemi Folusho, Shamsuddeen Hassan Muhammad, Ruqayya Nasir Iro and Maryam Sabo Abubakar for their assistance during data collection and annotation, thank you so much. David Adelani acknowledges the support of DeepMind Academic Fellowship programme. Finally, we thank the Spoken Language Systems Chair, Dietrich Klakow at Saarland University for providing GPU resources to train the models.

\bibliography{anthology,custom}
\bibliographystyle{acl_natbib}

\appendix

\section{Focus Languages}

\begin{table*}[!ht]
    \centering
    \resizebox{\textwidth}{!}{%
    \begin{tabular}{p{59mm}|p{60mm}|p{60mm}} 
    \toprule
         \textbf{English Translation} & \textbf{Target Language Translation} & \textbf{Literal Translation of Target language}  \\
        \midrule 
        \multicolumn{3}{l}{\textbf{Target Language: \yoruba}} \\ 
        \multicolumn{3}{l}{\textbf{Incorrect translation, sentiment not preserved.}} \\ 
        \midrule
        In the absence of such a perfect storm, avoid stabbing your wallet in the heart with this 'Dagger'. \colorbox{lightgray}{Definitely not recommended}  & N\'iw\`{o}n b\'{i} k`{o} ti s\'{i} `{i}j\`{i} l\'{i}le t\'{o} d\'{a}ra, m\'{a} \d{s}e fi "Dagger" y\`{i}\'{i} pa ow\'{o} r\d{e} n\'{i} \d{o}k\`{a}n r\d{e}. &  In the absence of a great storm, do not use this "Dagger" to kill your money in the heart \\ 
        \midrule
         \multicolumn{3}{l}{\textbf{Incorrect translation, sentiment preserved.}} \\ 
        \midrule
        Citation the movie. Perfect Movie. \colorbox{lightgray}{Loved every second of the movie.} Wished it didn't end  & Mo f\d{\'{e}}r\`{a} gbogbo \`{i}\d{s}\d{\'{e}}j\'{u} t\'{i} mo fi \'{n} \d{s}e f\'{i}\`{i}m\`{u} n\'{a}\`{a}, mo f\d{\'{e}} kí \'{o} m\'{a}\`{a} par\'{i} &  \colorbox{lightgray}{I enjoyed every second that I used to} make this movie. Wished it did not end\\ 
         \multicolumn{3}{l}{\textbf{Incorrect and Incomplete translation, sentiment not preserved}} \\ 
        \midrule
        Funny Funny Funny. Oh mehn, this movie is super funny. if you are looking for a movie to lift your mood up then \colorbox{lightgray}{this is the right movie for you}. & Orinrinrinrinrinrin... & song....... (MT output is nonsensical) \\
        \bottomrule
        \multicolumn{3}{l}{\textbf{Target Language: Igbo}} \\ 
        \multicolumn{3}{l}{\textbf{Incorrect translation, sentiment not preserved.}} \\ 
        \midrule
        Fifty minutes is spent advertising a holiday resort in Lagos, Movie closes. Money down the drain. \colorbox{lightgray}{Not recommended.}  & \d{O} b\d{u}r\d{u} na \d{i} na-eme ihe nd\d{i} a, \d{i} ga-enwe ike \d{i}hap\d{u} ya.& Do these things to leave it\\ 
        \midrule
         \multicolumn{3}{l}{\textbf{Incorrect translation, sentiment preserved.}} \\ 
        \midrule
         Temi Otedola's performance was truly stunning. \colorbox{lightgray}{I thoroughly enjoyed} the layers that the story had and the way that each key piece of information was revealed. & Ihe a o mere t\d{o}r\d{o} m ezigbo \d{u}t\d{o}, \d{o} na-at\d{o}kwa m \d{u}t\d{o} otú e si k\d{o}waa ihe nd\d{i} d\d{i} mkpa. & \colorbox{lightgray}{I thoroughly enjoyed} the layers that the story had and the way that each key piece of information was revealed.\\
         \midrule
         \multicolumn{3}{l}{\textbf{Incorrect and Incomplete translation, sentiment not preserved}} \\ 
        \midrule
        \colorbox{lightgray}{Nice cross-country movie.} The only thing that I don't like about this movie is the way there was little or no interaction with the Nigerian or Indian environment. \colorbox{lightgray}{Beautiful romantic movie}.  & Ihe m na-ad\d{i}gh\d{i} amas\d{i} na fim a b\d{u} na \d{o} d\d{i}gh\d{i} ihe jik\d{o}r\d{o} ya na nd\d{i} Na\d{i}jir\d{i}a ma \d{o} b\d{u} nd\d{i} India. & The only thing that I don't like about this movie is the way there was little or no interaction with the Nigerian or Indian environment\\
        \bottomrule
        \multicolumn{3}{l}{\textbf{Target Language: PCM - Nigerian Pidgin}} \\ 
        \midrule
        \multicolumn{3}{l}{\textbf{Incorrect translation, sentiment preserved.}} \\ 
        \midrule
        \colorbox{lightgray}{Nice cross-country movie}. The only thing that I don't like about this movie is the way there was little or no interaction with the Nigerian or Indian environment. \colorbox{lightgray}{Beautiful romantic movie}. & The only thing wey I no like about this film na because e no too get interaction with Nigerian or Indian people. &  The only thing that I don't like about this movie is the way there was little or no interaction with the Nigerian or Indian people.\\ 
        \midrule
         \multicolumn{3}{l}{\textbf{Incorrect translation, sentiment preserved.}} \\ 
        \midrule
         \colorbox{lightgray}{A flawed first feature film}, but it shows a great deal of promise  & Fear first feature film, but e show plenti promise. & Fear was featured in the film firstly but it shows a great deal of promise\\
          \midrule
         \multicolumn{3}{l}{\textbf{Incorrect and Incomplete translation, sentiment not preserved}} \\ 
        \midrule
        \colorbox{lightgray}{Spot On!!!} Definitely African movie of the year, enjoyed every minute of the 2hours 30minutes & Na almost every minute of the 2hours 30minutes wey dem take play for Africa film dem dey play. & It is almost every minute of the 2hours 30minutes that they play African movie they play\\
        \bottomrule
    \end{tabular}
    }
     \caption{\textbf{Examples of translation mistakes observed and impact on the sentiment}. The \colorbox{lightgray}{Gray color} identifies the sentiment portion of the review}

     \label{tab:qualitative_examples}
\end{table*}
\label{sec:focus_languages}
We focus on four Nigerian languages from three different language families. \textbf{Hausa} (hau) is from the Afro-Asiatic/Chadic family spoken by over 77 million (M) people.  \textbf{Igbo} (ibo) and \textbf{\yoruba} (yor) are both from Niger-Congo/ Volta-Niger family spoken by 30M and 46M respectively. While  \textbf{\naija} (pcm) is from the English Creole family, spoken by over 120M people. The \naija is ranked the 14th most spoken language in the world\footnote{https://www.ethnologue.com/guides/ethnologue200}. All languages make use of the Latin script. Except for \naija, the remaining are tonal languages. 
Also, Igbo and \yoruba make extensive use of diacritics in texts which are essential for the correct pronunciation of words and for reducing ambiguity in understanding their meanings.

\section{Hyper-parameters for PLMs}
\label{sec:parameters}
For fine-tuning PLMs, we make use of HuggingFace transformers~\citep{Wolf2019HuggingFacesTS}. We make use of maximum sequence length of $200$, batach size of $32$, number of epochs of $20$, and learning rate of $5e-5$ for all PLMs.

\section{Human Evaluation}

\label{sec:human_eval}
To verify the performance of the MT model, we hire at least two native speakers of each Nigerian indigenous languages - three native Igbo speakers, four native \yoruba speakers, four native speakers of Nigerian Pidgin and two Hausa native speakers. The annotators were individually given 100 randomly selected translated reviews in Excel sheets to report the adequacy and sentiment preservation (1: if they preserve sentiment, 0:otherwise) of the MT outputs. Alongside the sheets, the annotators are given an annotation guideline to guide them during the course of the annotation. Asides that the annotators are of the Nigerian descent as well as native speakers of the selected languages, their minimum educational experience is a bachelor's degree which qualifies them to efficiently read, write and comprehend the annotation materials and data to be annotated. 

To measure the consistency of our annotators, we added repeated 5 examples out of the 100 examples. Our annotators were consistent with their annotation. We measure the inter-agreement among the two annotators per task. For adequacy, the annotators achieved Krippendorff's alpha scores of 0.675, 0.443, 0.41, 0.65 for Hausa, Igbo, Nigerian-Pidgin, and \yoruba respectively. Similarly, for sentiment preservation, Krippendorff's alpha scores of 1.0, 0.93, 0.48, and 0.52 for Hausa, Igbo, Nigerian-Pidgin, and \yoruba respectively. In general, annotators reviewed the translated texts to have adequacy of 3.8 and 4.6. Nigerian-Pidgin (4.6) achieved better adequacy result as shown in \autoref{tab:automatic_metrics} because of her closeness to English language, Igbo was rated to have a lower adequacy score (3.8). Overall, all annotators rated the translated sentences to preserve sentiment at least in 90\% of the time i.e 90 out of 100 translations preserve the original sentiment in the English sentence. 

\subsection{Qualitative analysis}
The human evaluation is to verify the manually verify the quality of over 100 randomly selected translated sentences manually. Also, the reports from the annotators were automatically computed to support our claim that sentiment is usually preserved in MT outputs. The examples listed in \autoref{tab:qualitative_examples} are extracted during the annotation process. The examples illustrate the noticeable mistakes in MT outputs. The annotators are expected to give a rating scale between 1-5 if the randomly selected machine translated review is adequately translated and a binary 0-1 rating scale if the sentiment of the original review is retained in the the randomly selected machine translated review.

The examples that are listed in \autoref{tab:qualitative_examples} buttress our claim that MT outputs are not completely accurate as some translations in the target languages are missing thereby affecting the complete idea and meaning of the movie review that is originally written in English, which eventually could lead to losing the sentiment of the movie review. Also, as shown in \autoref{tab:qualitative_examples}, the sentiments of some reviews are preserved regardless of the incorrect or missing translations and the idea or meaning of the review is not totally lost.

\subsection{Annotation Guideline}

We provide the annotation guideline on Github\footnote{\url{}}. 

\end{document}